\documentclass[11pt]{article}

\usepackage[final]{acl}

\usepackage{times}
\usepackage{latexsym}
\usepackage{amsmath}
\usepackage{amssymb}

\usepackage{booktabs}

\usepackage[T2A, T1]{fontenc}

\usepackage[utf8]{inputenc}

\usepackage{microtype}

\usepackage{inconsolata}

\usepackage{graphicx}

\usepackage{listings}
\usepackage{xcolor}
\usepackage{colortbl}

\colorlet{punct}{red!60!black}
\definecolor{background}{HTML}{F8F8F8}
\definecolor{delim}{RGB}{20,105,176}
\colorlet{numb}{magenta!60!black}

\lstdefinelanguage{json}{
    basicstyle=\small\ttfamily,
    backgroundcolor=\color{background},
    showstringspaces=false,
    breaklines=true,
    frame=lines,
    literate=
     *{0}{{{\color{numb}0}}}{1}
      {1}{{{\color{numb}1}}}{1}
      {2}{{{\color{numb}2}}}{1}
      {3}{{{\color{numb}3}}}{1}
      {4}{{{\color{numb}4}}}{1}
      {5}{{{\color{numb}5}}}{1}
      {6}{{{\color{numb}6}}}{1}
      {7}{{{\color{numb}7}}}{1}
      {8}{{{\color{numb}8}}}{1}
      {9}{{{\color{numb}9}}}{1}
      {:}{{{\color{punct}{:}}}}{1}
      {,}{{{\color{punct}{,}}}}{1}
      {\{}{{{\color{delim}{\{}}}}{1}
      {\}}{{{\color{delim}{\}}}}}{1}
      {[}{{{\color{delim}{[}}}}{1}
      {]}{{{\color{delim}{]}}}}{1},
}

\definecolor{llmrow}{gray}{0.92}

\usepackage{tikz}
\usetikzlibrary{arrows.meta}      
\usetikzlibrary{shapes.geometric} 
\usetikzlibrary{shadows}          
\usetikzlibrary{fit}              
\usetikzlibrary{backgrounds}      

\title{LLMs as Feature Engineers for Text-and-Tabular Prediction}

\author{Merwan Barlier \\
  Teads\\
  \texttt{merwan.barlier@teads.com} \\\And
  Blaz Skrlj \\
  Teads\\
  \texttt{blaz.skrlj@teads.com} \\}

\begin{document}
\maketitle
\begin{abstract}
We introduce an iterative framework that automates the extraction of interpretable, schema-bound categorical features from unstructured text for tabular prediction models. To navigate the feature space, a generator LLM proposes semantic definitions, a separate extractor LLM materializes the features, and a downstream tabular model evaluates their predictive performance. We optimize this search by translating explicit model errors, such as AUC ranking inversions, into natural-language feedback, steering the LLM to resolve specific predictive failures. Evaluated across three public datasets, this error-driven loop accelerates feature discovery by up to $3\times$ compared to unguided search. Empirically, the generated features demonstrate strong multi-view complementarity, strictly outperforming any subset when combined with TF-IDF and dense embeddings. Finally, the framework guarantees instance-level interpretability: the discovered features dominate SHAP importance rankings and provide a fully transparent, semantic audit trail for every prediction.
\end{abstract}

\section{Introduction}

Tabular models routinely combine engineered numerical and categorical features but fundamentally struggle to incorporate raw text \citep{shi2021benchmarking}, a ubiquitous challenge in recommendation engines and click-through rate (CTR) prediction, where structured user metadata must seamlessly interact with unstructured item descriptions. Current approaches force a strict dichotomy: dense sentence-transformer embeddings \citep{reimers2019sentence} provide opaque predictive power without per-prediction explanations or human-readable structure, whereas hand-crafted features offer high interpretability but are computationally expensive to design and rarely transfer across tasks.

Large Language Models (LLMs) present a compelling alternative. Repurposed as zero-shot classifiers, LLMs can extract structured, high-level semantic attributes from raw text---mapping unstructured paragraphs into discrete categorical buckets such as a product's target demographic, a headline's emotional appeal, or an item's novelty \citep{Wangetal2023}. This allows for the generation of features that are simultaneously interpretable and highly predictive. However, discovering the optimal semantic dimensions for a specific task without laborious hand-curation remains an open challenge.

To address this, we introduce an iterative, agentic framework that automates the extraction of structured features from raw text. Within this loop, a \textit{generator} LLM proposes categorical feature definitions (comprising a name, a discrete value set, and a description), an \textit{extractor} LLM annotates the dataset by applying these definitions, and a downstream model evaluates the resulting features. This architecture is strictly task-agnostic and can optimize any target objective (e.g., classification AUC, regression MSE). It requires only two components: (1) a downstream tabular model to score candidates, and (2) a mechanism to translate concrete model errors into natural-language feedback to guide the generator LLM in the next iteration.

Our central methodological finding is that standard, unguided LLM proposals or simple scalar-score feedback loops fall short of discovering optimal features efficiently. Instead, the LLM must be presented with concrete, natural-language examples of where the current model fails---such as explicit ranking inversions for an AUC objective. By prompting the LLM to generate features that specifically separate these error cases, the iterative loop produces a highly compact feature set that is strictly complementary to the existing baseline model, driving direct improvements in downstream predictive performance.

\paragraph{Contributions}
\begin{itemize}
    \item \textbf{An iterative LLM-as-feature-engineer loop for text + tabular tasks.} Prior LLM feature engineering work operates strictly on tabular columns. We extend this paradigm to extract discrete, semantic categoricals directly from raw unstructured text.
    \item \textbf{Multi-view complementarity.} We confirm empirically across three public datasets of varying text complexity (Kickstarter, Amazon Books, Stack Overflow) that LLM-generated features, dense sentence embeddings, and classical TF-IDF capture conditionally independent signals. Combining all three text representations strictly outperforms any subset.
    \item \textbf{Instance-level interpretability.} Unlike dense embeddings, our discovered LLM features dominate downstream SHAP importance rankings and provide a transparent, auditable semantic trace for every individual prediction.
    \item \textbf{An error-driven textual feedback framework.} We demonstrate that translating downstream model errors (e.g., misranked pairs) into natural-language constraints effectively steers LLM feature generation, significantly accelerating convergence compared to an unguided search.
\end{itemize}

\section{Related Work}

\paragraph{LLMs as feature generators}
Recent work leverages LLMs for feature generation, but prior methods operate exclusively on structured tabular inputs—producing indicator rules \citep{choi2024featllm}, Python transformations \citep{hollmann2023caafe, ko2025fergllm}, decision-tree features \citep{nam2024octree}, or evolutionary tabular search \citep{abhyankar2025llmfe}. Crucially, none handle unstructured raw text. While \citet{abhyankar2025llmfe} similarly observe that structured feedback outperforms scalar scores, their scope remains strictly tabular. Furthermore, while Summary Boosting \citep{manikandan2023summaryboosting} flattens structured tabular attributes into natural-language strings, it forces the LLM to act as the final estimator, obligating a generative model to evaluate complex numerical thresholds within a prompt. Conversely, our framework operates in the exact opposite direction: it leverages the LLM strictly at design time to map unstructured text into discrete, schema-bound categories. By converting text into tabular features rather than tabular features into text, we enable a traditional classifier to natively optimize continuous numerical boundaries, delegating mathematical evaluation to the model framework best suited for it.

\paragraph{LLMs as evolutionary optimizers}
A growing literature leverages LLMs as variation operators to iteratively optimize structured artifacts. For instance, \emph{Evolution Through Large Models} \citep{lehman2022elm} and \emph{FunSearch} \citep{romeraparedes2024funsearch} introduced this paradigm for generating code and solving combinatorial problems. This approach has since been expanded to optimize prompts \citep{guo2024evoprompt, fernando2023promptbreeder} and discover reward functions \citep{ma2024eureka}. Closely related is \emph{OPRO} \citep{yang2024llmoptimizer}, which formalizes how an LLM can propose next-step candidates by reviewing a history of past scores. 

Our iterative loop builds upon this foundational paradigm, adapting it for feature engineering through two specific design choices. First, we focus the search space on \emph{schema-bound feature definitions} rather than free-form code. This ensures the outputs can be reliably applied as zero-shot classifiers while preserving SHAP-level interpretability. Second, instead of relying solely on scalar fitness scores, we provide the LLM with \emph{structured, natural-language descriptions of model errors} (e.g., AUC ranking inversions). By explicitly highlighting \emph{where} and \emph{why} the current model is struggling, this error-driven guidance helps the LLM navigate the feature space more efficiently, significantly accelerating convergence over an unguided search. Table~\ref{tab:conceptual_comparison} summarizes these structural distinctions against prior optimization and feature engineering frameworks.

\begin{table*}[t]
\centering
\small
\renewcommand{\arraystretch}{1.2}
\begin{tabular}{@{}p{2.5cm} p{2.8cm} p{3.2cm} p{3.8cm} p{2.2cm}@{}}
\toprule
\textbf{Method} & \textbf{Input Modality} & \textbf{Search Space} & \textbf{Feedback Signal} & \textbf{Live Inference} \\
\midrule
\textbf{CAAFE} \newline \citep{hollmann2023caafe} & Tabular only & Python code transformations & Zero-shot (No iterative feedback loop) & Fast (LLM-free) \\
\textbf{OPRO} \newline \citep{yang2024llmoptimizer} & Unstructured text & Free-form text prompts & Scalar metrics (Past fitness scores) & Slow (LLM call) \\
\midrule
\rowcolor{llmrow}
\textbf{Ours} & Raw text $\to$ Tabular & Schema-bound categorical definitions & Textual gradients (error-based) & Fast (LLM-free) \\
\bottomrule
\end{tabular}
\caption{Conceptual comparison of LLM-driven optimization and feature engineering frameworks. Unlike prior work, our approach bridges unstructured text and tabular prediction by iteratively optimizing schema-bound categories using explicit model failures, all while maintaining LLM-free real-time inference.}
\label{tab:conceptual_comparison}
\end{table*}
\section{Methodology}

\subsection{High-Level View}

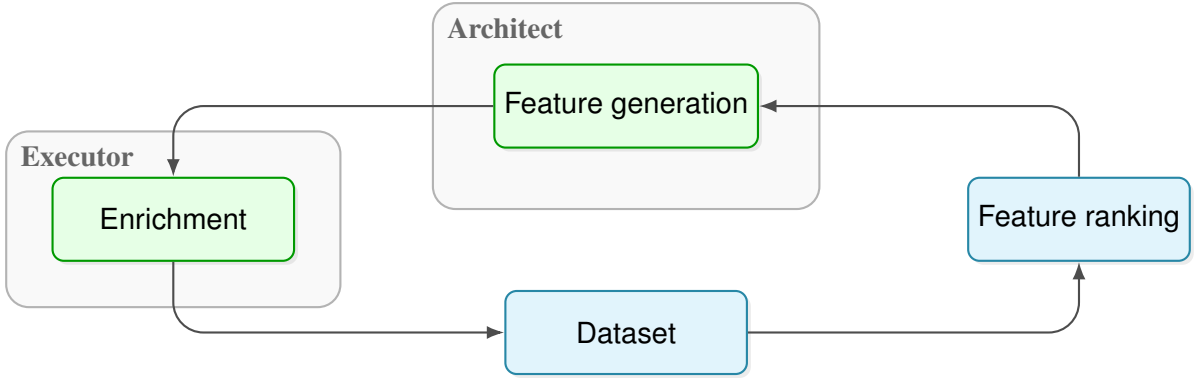
\begin{figure*}[t]
\centering
\resizebox{0.98\textwidth}{!}{%
\begin{tikzpicture}[
    >=Latex, 
    font=\sffamily,
    box/.style={rectangle, draw=black!60, thick, rounded corners=1.5mm, minimum height=1.1cm, minimum width=3.2cm, align=center, fill=white, drop shadow={opacity=0.15, shadow xshift=1.5pt, shadow yshift=-1.5pt}},
    diamondbox/.style={diamond, draw=orange!70!black, thick, aspect=2, minimum height=1.4cm, minimum width=3.5cm, align=center, fill=orange!10, drop shadow={opacity=0.15, shadow xshift=1.5pt, shadow yshift=-1.5pt}},
    greenbox/.style={box, fill=green!10, draw=green!60!black},
    bluebox/.style={box, fill=cyan!10, draw=cyan!60!black},
    architectbox/.style={rectangle, draw=black!30, thick, rounded corners=3mm, fill=gray!5},
    executorbox/.style={rectangle, draw=black!30, thick, rounded corners=3mm, fill=gray!5},
    filterbox/.style={rectangle, draw=red!60, dashed, thick, rounded corners=2mm, fill=red!10},
    line/.style={->, draw=black!70, thick, rounded corners=4mm}
]


\node[bluebox] (data) at (0, 0) {Dataset};
\node[greenbox] (enrich) at (-6, 1.5) {Enrichment};

\node[greenbox] (gen) at (0, 3) {Feature generation};
\node[bluebox, minimum width=2.6cm] (rank) at (6, 1.5) {Feature ranking};

\begin{scope}[on background layer]
    
    \node[architectbox, fit=(gen), inner xsep=8mm, inner ysep=8mm, 
          label={[anchor=north west, font=\bfseries, text=black!60, xshift=2pt, yshift=-2pt]north west:Architect}] (architect) {};

    \node[executorbox, fit=(enrich), inner xsep=6mm, inner ysep=6mm, 
          label={[anchor=north west, font=\bfseries, text=black!60, xshift=2pt, yshift=-2pt]north west:Executor}] (executor) {};

\end{scope}


\draw[line] (gen.west) -| (enrich.north);
\draw[line] (enrich.south) |- (data.west);
\draw[line] (data.east) -| (rank.south);  

\draw[line] (rank.north) |- (gen.east);

\end{tikzpicture}}
\caption{Agentic recommender workflow. An architect proposes features, an executor evaluates them, and feedback drives the search loop.}
\label{fig:flow}
\end{figure*}

Figure~\ref{fig:flow} illustrates our iterative agentic framework. At each step, a generator LLM proposes new feature definitions, an extractor materializes them across the dataset, a downstream tabular model evaluates their predictive utility, and the resulting errors are translated into natural-language feedback. Through this continuous loop, the LLM learns to propose increasingly effective features.

The system architecture is governed by three core design choices, each enforcing a critical property of the final feature set:

\begin{itemize}
    \item \textbf{Schema-Bound Categorical Definitions.} Rather than generating dense embeddings or free-form code, the LLM is constrained to output structured categorical definitions comprising a short name, a finite value set, and a semantic description (detailed in Section 3.2). This constraint provides three immediate benefits:
    \begin{itemize}
        \item \textit{Interpretability:} Features possess human-readable names and discrete states, enabling direct SHAP attribution and qualitative error analysis.
        \item \textit{Reliable Extraction:} Zero-shot classification into a small set of discrete buckets is a highly robust capability of current LLMs, avoiding the instability of continuous value prediction or arbitrary code execution.
        \item \textit{Decoupled Compute:} A feature definition is designed once by a frontier model (e.g., \texttt{GPT-5.4}) and subsequently applied across all rows by a faster, cost-effective batch model (e.g., \texttt{GPT-4.1-nano}), drastically amortizing inference costs.
    \end{itemize}
    
    \item \textbf{Objective Evaluation via Downstream Models.} The LLM does not score its own proposals. Instead, every generated feature is materialized and rigorously evaluated by a fast tabular model (\texttt{HistGradientBoostingClassifier}). This grounds the search in a concrete task metric (e.g., AUC) rather than relying on the LLM's internal confidence, avoiding the well-documented pitfalls of LLM self-evaluation.
    
    \item \textbf{Feedback as the Optimization Lever.} The iterative architecture remains strictly task-agnostic; its optimization behavior is entirely dictated by the natural-language feedback presented to the generator LLM. As detailed in Section~\ref{sec:feedback}, swapping the feedback signal fundamentally shifts the system's output to serve distinct use cases.
\end{itemize}

\subsection{Features as definitions}

A feature definition is a structured JSON object:

\begin{lstlisting}[language=json]
{
  "name": "sem_review_depth",
  "values": [
    "very_detailed", 
    "moderate", 
    "surface", 
    "barely_substantive"
  ],
  "description": "How much actual content the review contains. very_detailed = multiple paragraphs with specific examples; surface = a few sentences with general reactions; ..."
}
\end{lstlisting}

The LLM emits these in batches (one prompt → 90 definitions). Each definition is
then applied row-by-row by a separate, cheaper enrichment LLM
(\texttt{GPT-4.1-nano}) that acts as a zero-shot classifier: given the
description, the row text, and the candidate values, it picks one value.
Enrichment is amortised over the dataset and runs in parallel batches; for 80K rows this takes ~30 minutes.

This separation matters: the generator LLM needs to reason about what dimensions
of the text might matter for the task — a slow, complex job. The extractor
LLM only needs to apply a single label given clear criteria — a fast, parallel
job. Using the same model for both would be wasteful.

\subsection{The Iterative Loop}

Structurally, our agentic loop builds on the Optimization by Prompting (OPRO) framework \citep{yang2024llmoptimizer}. However, we extend OPRO by replacing its standard scalar-score feedback with rich, error-aligned textual constraints. 

More precisely, at each iteration $t$, the framework executes the following sequence:

\begin{enumerate}
    \item \textbf{Generate:} The \textit{generator} LLM proposes ${\sim}90$ new categorical feature definitions. This generation is strictly conditioned on: (a) a ``hall of fame'' containing the current best-performing features (names, value sets, and descriptions), (b) a running memory of all previously evaluated features to prevent duplicates, and (c) the current feedback signal (Section~\ref{sec:feedback}).
    \item \textbf{Enrich:} The \textit{extractor} LLM acts as a zero-shot classifier, applying each newly proposed definition to the dataset to produce discrete categorical columns.
    \item \textbf{Score:} Each new column is individually evaluated against the downstream task metric. The entire historical pool of generated features remains available for the subsequent selection phase.
    \item \textbf{Select \& Evaluate:} A greedy forward selection algorithm over the full candidate pool identifies the top-$K$ feature set. This optimized set drives the current iteration's headline metric and updates the hall of fame for the subsequent generation step.
    \item \textbf{Compute Feedback:} Based on the current model's predictions, we compute the target-specific feedback signal (detailed in Section~\ref{sec:feedback}) and formulate it as natural-language guidance for the next iteration.
\end{enumerate}

The batch size of ${\sim}90$ definitions per iteration strikes a deliberate balance: it provides the greedy selector with sufficient diversity while remaining small enough that each round of targeted feedback tightly steers the subsequent generation step.

\subsection{The Feedback Signal}
\label{sec:feedback}

The feedback signal dictates the framework's optimization behavior. To drive strict predictive improvements, we must surface concrete instances of model failure. By translating the mathematical concept of ``where the model is wrong'' into structured natural language, we explicitly steer the LLM's feature proposals toward complementing the current model.

For a binary classification task optimized for AUC, an ``error'' is defined as a ranking inversion: a true positive ranked below a true negative. We sample these misranked pairs from the current model's cross-validated predictions and present them explicitly to the generator LLM as contrastive text:

\begin{quote}
\small
\textbf{SUCCEEDED} (but model predicted 35\%): \textit{``Smart Herb Garden---grow fresh herbs''}\\
\textbf{FAILED} (but model predicted 62\%): \textit{``Revolutionary concept changes everything''}\\
$\rightarrow$ \textbf{Task:} Design a feature that explicitly separates these.
\end{quote}

To prevent the LLM from destroying existing predictive signal, we also present well-ranked pairs as negative constraints (``the model already gets these right---do not duplicate this signal''). This loss-aligned feedback perfectly targets the downstream objective: every inversion the LLM successfully separates directly improves the target metric. 

This error-driven formulation generalizes across tasks. For instance for Mean Squared Error (MSE) regression, the system surfaces items with the highest squared residuals; for ranking (e.g., NDCG), it provides misordered list positions. The underlying framework remains strictly task-agnostic; only the natural-language presentation of the error changes. Finally, to ensure the LLM remains calibrated to the task scale, the prompt includes the current baseline metric, the total count of residual errors, and the marginal value of fixing a single instance.

\subsection{Feature Selection}

While the LLM acts as the generator of categorical definitions, the downstream model serves as the definitive evaluator. After the enrichment phase, we isolate the most complementary features from the generated pool using a greedy forward selection algorithm \cite{guyon2003introduction}.

The procedure initializes with a baseline model trained exclusively on the existing tabular features. At each step, every candidate feature in the accumulated pool is temporarily added to the active feature set, and the model is re-evaluated on the downstream target metric. The single candidate that yields the highest performance gain is permanently absorbed into the selected set. This incremental process repeats until no remaining candidate improves the metric, or until a predefined feature budget is exhausted.

\section{Experiments}

\subsection{Datasets and tasks}
\label{sec:datasets}

We evaluate our framework on three public text-and-tabular prediction tasks, deliberately selected to span a spectrum of unstructured text complexity---from short, dense titles to long, structured documents. To ensure consistent evaluation, all datasets are subsampled to a comparable scale (60K--80K rows) and split into stratified 5-fold cross-validation sets using a fixed random seed. Complete details regarding raw dataset sourcing, data filtering pipelines, and target variable binarization are provided in Appendix~\ref{app:datasets}.

\paragraph{Kickstarter (Short Text).} 80K crowdfunding projects sourced from Kaggle, filtered to completed projects (successful or failed). The unstructured text is the project name (averaging 3--12 words). Tabular features include main category, subcategory, currency, country, \texttt{goal\_usd}, \texttt{duration\_days}, \texttt{launch\_month}, \texttt{launch\_year}, and launch hour. The target is binary project success (baseline rate: ${\sim}40\%$). This represents the short-text regime, where the semantic signal is real but highly concentrated.

\paragraph{Amazon Books Reviews (Medium Text).} 80K book reviews sampled from the Kaggle Amazon dataset, restricted to reviews with at least 5 total helpfulness votes. The text is the review body itself (median length ${\sim}680$ characters). Tabular features include star rating, item price, review timestamp, and total helpfulness votes. The target is binary \texttt{helpful}, defined as a helpful-vote ratio of $\geq 0.7$ (baseline rate: ${\sim}60\%$). This represents the medium-text regime, featuring substantive, opinionated, but bounded paragraphs.

\paragraph{Stack Overflow Questions (Long, Structured Text).} 60K programming questions from the public Kaggle dataset assessing question quality. Originally human-labeled into three tiers, we binarize the target to High Quality (HQ) versus not-HQ (baseline rate: ${\sim}33\%$). The text concatenates the question title and body (median length ${\sim}780$ characters), inherently including complex structures like code blocks, error messages, and stack traces. Tabular features include the primary tag, number of tags, body length, title length, and temporal creation features (year, month, day, hour). This represents the text-richest regime, offering the largest information capacity per row.

\paragraph{The Text-Complexity Spectrum.} Together, these three datasets capture distinct regimes of text informativeness. Kickstarter relies on ultra-short hooks, Amazon Books provides conversational paragraphs, and Stack Overflow introduces long, multi-format technical questions. This progression explicitly allows us to evaluate how the LLM-as-feature-engineer framework scales as the unstructured text becomes increasingly central to the prediction task.

\subsection{Implementation Details}
\label{sec:implementation}

\paragraph{Models and Architecture.} Feature generation is handled by a frontier LLM (in our case \texttt{GPT-5.4}), which is invoked purely at design time to propose batches of approximately 90 new feature definitions per iteration. The subsequent row-by-row extraction is performed by a more cost-effective model (\texttt{GPT-4.1-nano}), acting as a zero-shot classifier over the text. For the downstream evaluator, we utilize a \texttt{HistGradientBoostingClassifier} from \texttt{scikit-learn} \cite{pedregosa2011scikit}, as it natively and efficiently handles discrete categorical variables without the need for high-dimensional one-hot encoding. For full reproducibility, the exact instruction templates used for both the generator LLM (incorporating the loss-aligned feedback) and the extractor LLM are reproduced in Appendix~\ref{app:gen_prompt} and Appendix~\ref{app:ext_prompt}, respectively.

\paragraph{Compute and Hyperparameters.} Evaluating the entire historical pool of generated features during greedy selection is computationally intensive but trivially parallelizable. We distribute candidate evaluation across 32 workers (via \texttt{joblib.Parallel}). Under this configuration, evaluating a pool of 1,500 candidate features takes approximately 5 minutes per greedy step. Across all experiments, we cap the greedy forward selection algorithm at a maximum feature budget of $K=10$.

\subsection{Prediction with LLM features}
\label{sec:auc}

We instantiate the loss-aligned variant of our framework (Section~3.4) and evaluate our LLM-generated categorical features against two standard text representations: classical TF-IDF (top-5000 unigrams and bigrams, reduced to 50 components via TruncatedSVD) and dense sentence embeddings (\texttt{all-MiniLM-L6-v2}, reduced to 50 components via PCA). All text representations are concatenated to the tabular baseline. We report 5-fold cross-validated AUC, with significance tested via paired $t$-tests across folds.

\begin{table*}[t]
\centering
\small
\setlength{\tabcolsep}{6pt}
\begin{tabular}{lccc}
\toprule
Method                          & Kickstarter         & Amazon Books        & Stack Overflow \\
\midrule
Tabular only                    & $0.7444$            & $0.7611$            & $0.8544$ \\
\midrule
+ TF-IDF SVD-50                 & $0.7528$            & $0.8248$            & $0.9468$ \\
+ MiniLM PCA-50                 & $0.7572$            & $0.8201$            & $\mathbf{0.9510}$ \\
+ 10 LLM features (ours)        & $\mathbf{0.7639}$   & $\mathbf{0.8366}$   & $0.9320$ \\
\midrule
+ LLM + TF-IDF                  & $0.7662$            & $0.8445$            & $0.9599$ \\
+ LLM + Embeddings              & $0.7680$            & $0.8481$            & $0.9612$ \\
\textbf{+ LLM + TF-IDF + Emb}   & $\mathbf{0.7692}$   & $\mathbf{0.8510}$   & $\mathbf{0.9700}$ \\
\bottomrule
\end{tabular}
\caption{Multi-view comparison: 5-fold cross-validated AUC for each combination of representations on each dataset. Bold values mark the best single-view text representation (top group) and the best overall combination (bottom row). LLM features are the strongest single view on Kickstarter and Amazon, while MiniLM is strongest on Stack Overflow --- yet adding LLM features always improves on top of either alternative, and all three combined is strictly best across datasets.}
\label{tab:multiview}
\end{table*}

\paragraph{Three-View Complementarity.} Table~\ref{tab:multiview} details the predictive performance across all feature combinations. Among the single-view text representations (top block), LLM features are the strongest standalone addition for Kickstarter and Amazon Books. On Stack Overflow, dense embeddings edge out LLM features—a shift consistent with the dataset's dense, code-heavy text structure.

Crucially, however, the three representations exhibit \emph{conditional independence}. Across all three datasets, adding LLM features to TF-IDF yields a strictly larger improvement than either baseline alone. Adding LLM features to dense embeddings provides an even greater boost, and combining all three views (lexical + distributional + semantic) achieves the highest overall AUC. This aligns perfectly with multi-view learning theory \citep{blum1998cotraining, sridharan2008mvinfo}, which posits that conditionally independent views of the same input jointly improve performance. Much like established paradigms combining TF-IDF with dense word embeddings in NLP, our LLM categorical features extract explicit semantic dimensions that remain invisible to both traditional token-based and dense vector representations.

\paragraph{Error-driven feedback dramatically accelerates convergence.}
To isolate the value of error-aligned feedback, we ran a rigorous no-feedback ablation (``random + memory'') utilizing the same generator LLM, the same ~90 features per iteration, and the same 15-iteration budget. In this baseline, the LLM only sees a hall of fame of past best features and a list of features not to duplicate, but receives no error-driven guidance. Table~\ref{tab:method_vs_baseline} reports the final $\Delta$AUC after 15 iterations for both variants, along with the specific iteration where our feedback-guided method eclipses the baseline's final ceiling.

\begin{table*}[t]
\centering
\small
\setlength{\tabcolsep}{16pt}
\begin{tabular}{lccc}
\toprule
Dataset       & Method                 & Baseline               & Reaches\\
              & $\Delta$AUC            & $\Delta$AUC            & at iter \\
\midrule
Kickstarter   & $\mathbf{+0.0225}$     & $+0.0212$              & 4 \\
Amazon Books  & $\mathbf{+0.0731}$     & $+0.0679$              & 3 \\
Stack Overflow& $\mathbf{+0.0761}$     & $+0.0752$              & 9 \\
\bottomrule
\end{tabular}
\caption{Method vs. no-feedback baseline. Both variants run for 15 iterations with identical generator/extractor models and per-iteration budgets. Our method strictly outperforms the baseline's final $\Delta$AUC across all three datasets. More importantly, the ``Reaches at iter'' column demonstrates that loss-aligned feedback achieves the baseline's absolute peak performance in a fraction of the time, delivering up to a 5x reduction in LLM inference costs for the same downstream performance.}
\label{tab:method_vs_baseline}
\end{table*}

While a brute-force random search with memory can eventually stumble upon useful features given enough iterations, our error-aligned feedback loop navigates the feature space with remarkable efficiency. Table~\ref{tab:method_vs_baseline} shows that our method strictly dominates the 15-iteration baseline across all three datasets, securing particularly notable absolute gains on Amazon Books (+0.0052 $\Delta$AUC). However, the most profound advantage of our approach lies in its convergence speed. By explicitly targeting model failures, the feedback-guided variant achieves the baseline's absolute peak performance in a fraction of the time---requiring only 3 iterations on Amazon Books (a ~5x reduction in LLM inference costs) and 4 iterations on Kickstarter. In industrial applications where LLM API budgets and design-time compute are primary bottlenecks, this ability to rapidly and intelligently converge on high-signal features makes error-driven guidance indispensable.

\subsection{Interpretability: SHAP analysis of selected features}
\label{sec:shap}

A core advantage of our framework is the semantic interpretability of LLM-generated features---a property dense sentence embeddings inherently lack. To substantiate this, we conduct a SHAP \citep{lundberg2017unified} analysis to explicitly quantify each feature's marginal contribution to the predictions. 

For each dataset, we train the downstream model using the tabular baseline augmented with the top 10 selected LLM features. Computing the mean absolute SHAP values across the test set reveals a striking result: the generated semantic features do not merely contribute positively; they \emph{dominate} global feature importance. Across all three domains, multiple LLM features rank strictly above the strongest hand-engineered tabular baselines.

\begin{itemize}
    \item \textbf{Kickstarter.} The most influential feature overall is the LLM-generated \texttt{sem\_specificity\_of\_core\_noun} (mean |SHAP| $= 2.25$), which classifies the semantic clarity of the project title. It is nearly 10$\times$ stronger than the second feature, \texttt{sem\_target\_audience} (0.24, also LLM-generated). By contrast, the dominant hand-engineered tabular feature, \texttt{goal\_usd}, ranks only sixth (0.09). In total, 7 of the top 15 features are LLM-generated, including the top two slots.
    
    \item \textbf{Amazon Books.} The same pattern holds. The top feature is the LLM-generated \texttt{sem\_content\_materiality\_focus} (0.59), which categorizes the conceptual versus physical nature of the review, followed by \texttt{sem\_negative\_target\_location} (0.45). The strongest tabular feature, \texttt{review\_time}, ranks third (0.35), and the user-provided \texttt{review\_score} is seventh (0.16). Overall, LLM-generated features occupy 10 of the top 14 slots.
    
    \item \textbf{Stack Overflow.} The pattern reaches its strongest form on the text-richest dataset. The single most important feature is the LLM-generated \texttt{struct\_title\_orthography\_quality} (1.10), which assesses capitalization and spelling, easily beating the strongest tabular feature, \texttt{primary\_tag} (0.67). Another LLM feature, \texttt{tone\_formality\_vs\_fragmentation} (0.40), ranks fourth. In total, 5 of the top 10 overall most-influential features are LLM-generated.
\end{itemize}

Table~\ref{tab:shap_top10} reports the top-10 features by mean $|\text{SHAP}|$. As indicated by the shaded cells, LLM-generated features occupy the top slot on every dataset and completely dominate the overall rankings. For qualitative assessment, the exact JSON definitions of the top three LLM features per dataset---including their complete value sets and natural-language descriptions---are provided in Appendix~\ref{app:topfeats}.

\begin{table*}[t]
\centering
\scriptsize
\setlength{\tabcolsep}{3pt}
\renewcommand{\arraystretch}{1.15}
\begin{tabular}{p{4.3cm}r @{\hskip 0.8em} p{4.3cm}r @{\hskip 0.8em} p{4.3cm}r}
\toprule
\multicolumn{2}{c}{\textbf{Kickstarter}} & \multicolumn{2}{c}{\textbf{Amazon Books}} & \multicolumn{2}{c}{\textbf{Stack Overflow}} \\
\midrule
\cellcolor{llmrow}\texttt{sem\_specificity\_of\_core\_noun} & \cellcolor{llmrow}2.25 & \cellcolor{llmrow}\texttt{sem\_content\_materiality\_focus} & \cellcolor{llmrow}0.59 & \cellcolor{llmrow}\texttt{struct\_title\_orthography\_quality} & \cellcolor{llmrow}1.10 \\
\cellcolor{llmrow}\texttt{sem\_target\_audience} & \cellcolor{llmrow}0.24 & \cellcolor{llmrow}\texttt{sem\_negative\_target\_location} & \cellcolor{llmrow}0.45 & \texttt{primary\_tag} & 0.67 \\
\texttt{country} & 0.17 & \texttt{review\_time} & 0.35 & \texttt{creation\_year} & 0.62 \\
\texttt{launch\_year} & 0.15 & \cellcolor{llmrow}\texttt{sem\_inspection\_depth} & \cellcolor{llmrow}0.30 & \cellcolor{llmrow}\texttt{tone\_formality\_vs\_fragmentation} & \cellcolor{llmrow}0.40 \\
\texttt{category} & 0.15 & \cellcolor{llmrow}\texttt{sem\_review\_depth} & \cellcolor{llmrow}0.24 & \texttt{n\_tags} & 0.37 \\
\texttt{goal\_usd} & 0.09 & \cellcolor{llmrow}\texttt{sem\_bias\_or\_agenda\_signal} & \cellcolor{llmrow}0.19 & \cellcolor{llmrow}\texttt{struct\_body\_formatting\_stability} & \cellcolor{llmrow}0.32 \\
\cellcolor{llmrow}\texttt{sem\_title\_community\_reference\_quality} & \cellcolor{llmrow}0.08 & \texttt{review\_score} & 0.16 & \texttt{creation\_hour} & 0.31 \\
\cellcolor{llmrow}\texttt{sem\_release\_context\_specificity\_refined} & \cellcolor{llmrow}0.07 & \cellcolor{llmrow}\texttt{sem\_textual\_anchor\_type} & \cellcolor{llmrow}0.12 & \cellcolor{llmrow}\texttt{sem\_environment\_anchor\_specificity} & \cellcolor{llmrow}0.27 \\
\texttt{main\_category} & 0.05 & \cellcolor{llmrow}\texttt{sem\_review\_subject\_domain} & \cellcolor{llmrow}0.10 & \texttt{creation\_month} & 0.19 \\
\texttt{duration\_days} & 0.04 & \cellcolor{llmrow}\texttt{struct\_revision\_or\_editing\_signal} & \cellcolor{llmrow}0.10 & \cellcolor{llmrow}\texttt{sem\_framework\_magic\_dependency} & \cellcolor{llmrow}0.18 \\
\bottomrule
\end{tabular}
\caption{Top-10 features by mean $|$SHAP$|$ across the test set for each dataset. LLM-generated features are highlighted (shaded cells). The top feature on every dataset is LLM-generated, and the LLM-feature dominance strengthens as the text content of the task gets richer.}
\label{tab:shap_top10}
\end{table*}

\paragraph{Instance-Level Interpretability.} Beyond global trends, our framework guarantees per-prediction interpretability. Because LLM-generated features map to human-readable concepts, SHAP can decompose individual decisions into transparent audit trails. For example, a confident ``helpful'' Amazon review prediction traces to exact semantic factors: engaging with concepts (\texttt{sem\_content\_\allowbreak materiality\_\allowbreak focus=ideas\_\allowbreak arguments}, $+0.36$), deep analysis (\texttt{sem\_inspection\_depth=deep}, $+0.18$), and a 5-star rating ($+0.09$). This granular transparency is fundamentally impossible with dense embeddings, where SHAP only attributes importance to opaque vector dimensions (e.g., ``Dimension 14''). By making every prediction fully auditable, our approach is ideal for deployments requiring algorithmic verification by domain experts.

\section{Industrial Deployment and Performance Trade-offs}

To validate the real-world utility of agentic feature discovery, the framework was deployed across two live Click-Through Rate (CTR) prediction models in a high-throughput recommendation system \cite{mcmahan2013ad}. Offline evaluation demonstrated modest but consistent predictive improvements, yielding relative Information Gain (RIG) lifts of approximately +0.1\% to +0.26\%. However, online A/B testing revealed that these offline metrics understated the true business impact once the models interacted with live bidding dynamics and cold-start traffic. In live traffic experiments, the LLM-augmented models achieved substantial financial gains. One traffic deployment yielded a +6.6\% increase in Gross Revenue (GR) and a +6.3\% lift in a monitored margin proxy. A second deployment drove a +6\% margin proxy lift and a +5\% improvement in Conversion-Value Utilization (CVU), supported by cold-start AUC wins in 12 out of 13 early-life hour buckets.

Crucially, these financial and predictive gains required material computational trade-offs. The inclusion of these LLM-generated categorical features and their resulting feature crosses substantially increased serving complexity. Across the two deployments, inference latency rose by 30\% to 50\%, and CPU time per ad increased by 25\% to 30\%. In one traffic segment, this latency overhead resulted in a +6.5\% increase in critical timeout losses. Consequently, the final production deployments necessitated strict real-time monitoring of client error rates, underscoring that future agentic loops must explicitly optimize for strict computational budgets alongside predictive relevance.

\section{Conclusion}

We introduced an agentic framework that extracts schema-bound categorical features from unstructured text to enhance tabular models. We demonstrated that translating downstream ranking inversions into natural-language feedback explicitly guides the LLM through the feature space, accelerating convergence over unguided search. 

Significantly, these LLM-extracted features and traditional dense embeddings capture orthogonal predictive signals. While embeddings map continuous distributional semantics, our schema-bound features extract discrete conceptual dimensions. Combining these paradigms strictly outperforms either approach in isolation. Beyond predictive gains, these discrete features resolve the interpretability bottleneck of dense embeddings, enabling SHAP to generate transparent semantic audit trails for every individual prediction. This orthogonal complementarity suggests agentic feature discovery is a highly promising direction for unlocking the full value of unstructured text. Future work will extend this framework to multimodal inputs, continuous targets, and open-weights models.

\newpage

\section*{Limitations}

While our framework effectively bridges unstructured text and tabular prediction, it faces three primary limitations. First, relying on proprietary models (e.g., \texttt{GPT-5.4}) introduces API dependencies and limits exact reproducibility. Future work should evaluate whether open-weights alternatives (e.g., Llama 3) can reliably execute this error-driven search. Second, our evaluation is restricted to English datasets. It remains unclear how well the generator's semantic priors transfer to low-resource languages or specialized, jargon-heavy domains without fine-tuning. Finally, although LLM computation occurs offline, adding new categorical features still impacts live serving. In our industrial deployment, expanded feature vectors increased inference latency by 30\% to 50\%. In latency-critical environments, practitioners must balance predictive gains against computational costs, potentially by restricting the maximum feature budget.

\section*{Ethical Considerations}

While our framework improves predictive performance and interpretability, its deployment introduces three key ethical risks:

\begin{itemize}
    \item \textbf{Propagation of Bias:} Generator and extractor LLMs can introduce historical, cultural, or linguistic biases from their training data into the engineered semantic features, which the downstream model may subsequently amplify.
    \item \textbf{Metric Over-Optimization:} Optimizing loops strictly for downstream performance or commercial metrics, such as our observed $+6.6\%$ Gross Revenue lift, can inadvertently incentivize polarizing content. We actively mitigate this risk through instance-level SHAP interpretability, which provides a transparent semantic audit trail enabling human-in-the-loop verification and moderation.
    \item \textbf{Data Privacy:} Processing raw, user-generated text fields that may contain personally identifiable information (PII) introduces privacy vulnerabilities during API-based extraction, necessitating strict data-masking and anonymization preprocessing.
\end{itemize}

\section*{AI Assistant Acknowledgment}
During the preparation of this work, the authors utilized Large Language Models (LLMs) to assist with brainstorming experimental designs, generating and debugging code for the empirical pipeline, and refining the prose of the manuscript. All AI-assisted outputs were rigorously reviewed, edited, and verified by the human authors, who assume full responsibility for the final contents and claims of this paper.


\newpage
\bibliography{custom}

\newpage
\appendix

\section{Dataset construction}
\label{app:datasets}

All three datasets are derived from public Kaggle releases and processed
into a uniform schema: a single text column, a small tabular metadata
table, and a binary target. Subsampling and stratification use a fixed
random seed (\texttt{42}) for full reproducibility.

\subsection{Kickstarter}
\textbf{Source:} \texttt{kemical/kickstarter-projects} on Kaggle
(\texttt{ks-projects-201801.csv}, 378K rows). \textbf{Filtering:} we keep
only projects whose final state is \texttt{successful} or \texttt{failed}
(dropping \texttt{live}, \texttt{canceled}, \texttt{suspended}, and
\texttt{undefined}), then uniformly downsample to 80K rows, preserving the
original class balance ($\sim$40\% successful). \textbf{Text column:}
\texttt{name} (the project title, 3--12 words). \textbf{Tabular columns:}
\texttt{main\_category}, \texttt{category}, \texttt{currency},
\texttt{country}, \texttt{goal} (converted to USD as \texttt{goal\_usd}),
\texttt{deadline} and \texttt{launched} converted to
\texttt{duration\_days}, \texttt{launch\_year}, \texttt{launch\_month},
\texttt{launch\_dayofweek}, \texttt{launch\_hour}. \textbf{Label:}
\texttt{success} $= 1$ if final state is \texttt{successful}, else $0$.

\subsection{Amazon Books reviews}
\textbf{Source:} the Amazon Books Reviews dataset on Kaggle
(\texttt{Books\_rating.csv}, 3M reviews). \textbf{Filtering:} we restrict
to reviews with at least 5 total helpfulness votes (\texttt{helpful\_total}
$\geq 5$) to avoid noisy single-vote signals, then uniformly downsample to
80K rows. \textbf{Text column:} \texttt{review/text} (median length
${\sim}682$ characters). \textbf{Tabular columns:} \texttt{review/score}
(1--5 stars), \texttt{Price}, \texttt{review/time} (Unix timestamp,
converted to years since 2000), \texttt{helpful\_total} (total votes).
\textbf{Label:} \texttt{helpful} $= 1$ if helpfulness ratio
\texttt{helpful\_positive / helpful\_total} $\geq 0.7$, else $0$
(baseline rate ${\sim}60\%$).

\subsection{Stack Overflow questions}
\textbf{Source:} \textbf{Source:} \texttt{stackoverflow/\allowbreak 60k-stack-overflow-\allowbreak questions-with-\allowbreak quality-rate}
on Kaggle (60K rows, three quality tiers). \textbf{Label binarization:} we
collapse the three original labels (\texttt{HQ}, \texttt{LQ\_EDIT},
\texttt{LQ\_CLOSE}) into a binary target: \texttt{high\_quality} $= 1$ if
\texttt{HQ}, else $0$ (baseline rate ${\sim}33\%$). \textbf{Text column:}
\texttt{text} formed by concatenating \texttt{Title} and \texttt{Body} with
\texttt{"|||"} as separator (median length ${\sim}780$ characters,
containing HTML, code blocks, error messages, and stack traces).
\textbf{Tabular columns:} \texttt{primary\_tag} (first tag in the
\texttt{Tags} field), \texttt{n\_tags} (number of tags),
\texttt{body\_length} and \texttt{title\_length} (character counts), and
temporal features \texttt{creation\_year}, \texttt{creation\_month},
\texttt{creation\_dayofweek}, \texttt{creation\_hour} parsed from
\texttt{CreationDate}.

\section{Generator prompt}
\label{app:gen_prompt}

The generator LLM receives a single prompt assembling six sections in
order: a task instruction, a diversity requirement, the current
performance baseline, available strategies, the hall of fame, the
feedback signal, and the output format. We give the full template below
with $\langle$angled placeholders$\rangle$ that are filled in per
iteration.

\begin{lstlisting}[basicstyle=\scriptsize\ttfamily,breaklines=true,breakatwhitespace=true]
You are a feature engineering expert. Your ONLY
job: output a JSON object with feature definitions.
Do NOT explain your reasoning. Do NOT output
anything except valid JSON.

Generate <N> features extracted from text column
<TEXT_COL> that improve the model's ability to
rank items correctly (predict which will succeed
vs fail).

## Current model performance

  Baseline AUC (tabular only): <BASE_AUC>
  Total misranked pairs to fix: <N_MISRANKED>

Each feature you propose will be scored by how
much it improves this metric.

## Where the model ranks WRONG

These pairs are ranked incorrectly --- the model
thinks the negative is more likely positive than
the positive:

  SUCCEEDED (pred=<P_succ>%): "<text_succ>"
  FAILED    (pred=<P_fail>%): "<text_fail>"
  -> Design a feature that separates these.

[ ... 5 misranked pairs in total ... ]

## Where the model ranks CORRECTLY (do not duplicate)

  SUCCEEDED (pred=<P_succ>%): "<text_succ>"
  FAILED    (pred=<P_fail>%): "<text_fail>"

[ ... 3 well-ranked pairs ... ]

## Current best features
<TOP_K HoF features, each with name, type, values,
description, and ScuiAUC contribution>

## Already tested (do NOT re-propose)
<comma-separated list of all past feature names>

## Output format

Output ONLY a valid JSON object. No markdown, no
explanation. Each feature must have: name
(snake_case with domain prefix like sem_ or
struct_), type ("single"), values (list of 3-10
strings), description (string).
\end{lstlisting}

\section{Extractor prompt}
\label{app:ext_prompt}

The extractor LLM receives one prompt per (row, feature) pair via the
batch API. The prompt is generated automatically from the feature
definition by the \texttt{llm-enrichment-lib} library; we reproduce its
structure here.

\begin{lstlisting}[basicstyle=\scriptsize\ttfamily,breaklines=true,breakatwhitespace=true]
You are a structured-data extractor. Given the
input text, classify it into one of the listed
values for each feature. Output ONLY a valid JSON
object mapping feature name to "value;confidence"
where confidence is in [0, 1].

## Input text
<text_column_value>

## Features to extract

### <feature_name>
<feature_description>
Values: <comma-separated list of allowed values>

[ ... one block per feature in the current batch ... ]

## Output format

A single JSON object:
{
  "<feature_name_1>": "<value>;<confidence>",
  "<feature_name_2>": "<value>;<confidence>",
  ...
}
\end{lstlisting}

In post-processing, the confidence suffix is stripped from the value
(\texttt{"clear;0.9"} $\to$ \texttt{"clear"}), and the resulting
categorical column is appended to the working CSV.

\section{Top discovered features by dataset}
\label{app:topfeats}

For each dataset we list the three LLM-generated features with the
highest mean $|$SHAP$|$ (from Table~\ref{tab:shap_top10}), reproducing
the exact name, type, value set, and natural-language description as
proposed by the generator LLM. These are the actual features used by
the final downstream model, unedited.

\subsection{Kickstarter}

\paragraph{1. \texttt{sem\_specificity\_of\_core\_noun}} \emph{(single)}\\
Values: \textit{highly\_specific\_core\_noun}, \textit{moderately\_specific\_core\_noun},
\textit{broad\_category\_noun}, \textit{vague\_abstract\_core\_noun},
\textit{no\_clear\_core\_noun}, \textit{unclear}.\\
Description: ``Specificity of the main noun anchoring the title, such as
dictionary, guide, album, ball, economy, or art.''

\paragraph{2. \texttt{sem\_target\_audience}} \emph{(single)}\\
Values: \textit{mass\_market}, \textit{niche\_hobby}, \textit{tech\_enthusiasts},
\textit{artists\_creators}, \textit{families}, \textit{activists}, \textit{unclear}.\\
Description: ``Who the project targets. \textit{mass\_market} = broad appeal;
\textit{niche\_hobby} = specific interest group; \textit{tech\_enthusiasts}
= gadget/software people; \textit{artists\_creators} = creative community;
\textit{families} = children/parents; \textit{activists} = social/political
cause; \textit{unclear} = no clear target.''

\paragraph{3. \texttt{sem\_title\_community\_reference\_quality}} \emph{(single)}\\
Values: \textit{community\_with\_specific\_project}, \textit{community\_as\_beneficiary},
\textit{community\_as\_identity\_signal}, \textit{community\_without\_project\_clarity},
\textit{no\_community\_reference}, \textit{unclear}.\\
Description: ``How community references are used and whether they clarify
the project.''

\subsection{Amazon Books}

\paragraph{1. \texttt{sem\_content\_materiality\_focus}} \emph{(single)}\\
Values: \textit{abstract\_impressions}, \textit{ideas\_arguments},
\textit{narrative\_events}, \textit{examples\_exercises},
\textit{material\_object\_quality}, \textit{mixed\_materiality}.\\
Description: ``What kind of material the reviewer is chiefly engaging:
abstract impressions, ideas, narrative events, examples/exercises, or
physical object quality.''

\paragraph{2. \texttt{sem\_negative\_target\_location}} \emph{(single)}\\
Values: \textit{none}, \textit{book\_content}, \textit{author\_claims},
\textit{edition\_production}, \textit{publisher\_packaging},
\textit{other\_reviewers\_or\_public}, \textit{mixed\_targets}.\\
Description: ``Where criticism is aimed: the book's content, the author's
claims, the edition/production, packaging/publisher framing, other
reviewers/public discourse, or mixed.''

\paragraph{3. \texttt{sem\_inspection\_depth}} \emph{(single)}\\
Values: \textit{glance\_level}, \textit{sampled\_portions},
\textit{substantial\_portions}, \textit{close\_inspection},
\textit{systematic\_inspection}.\\
Description: ``Apparent depth of inspection of the book or edition
reflected in the review.''

\subsection{Stack Overflow}

\paragraph{1. \texttt{struct\_title\_orthography\_quality}} \emph{(single)}\\
Values: \textit{professional}, \textit{minor\_issues},
\textit{casual\_lowercase}, \textit{noisy\_typos}, \textit{chaotic\_stylized}.\\
Description: ``Overall orthographic quality of the title, capturing
capitalization, spelling, and stylized noisy writing.''

\paragraph{2. \texttt{tone\_formality\_vs\_fragmentation}} \emph{(single)}\\
Values: \textit{formal\_complete}, \textit{neutral\_complete},
\textit{informal\_complete}, \textit{telegraphic\_fragmented},
\textit{chaotic\_fragmented}.\\
Description: ``Joint view of tone formality and sentence completeness.''

\paragraph{3. \texttt{struct\_body\_formatting\_stability}} \emph{(single)}\\
Values: \textit{stable\_and\_readable}, \textit{minor\_breaks},
\textit{noticeably\_irregular}, \textit{unstable\_or\_messy}.\\
Description: ``Visual stability of body formatting including paragraphs,
line breaks, and transitions.''

\end{document}